 \documentclass[final,3p,times,twocolumn,authoryear]{elsarticle}

\usepackage{amssymb}
\usepackage{lipsum}

\usepackage{bibentry}

\usepackage{algorithm}
\usepackage{algorithmic}

\usepackage{caption}
\usepackage{subcaption}
\usepackage[utf8]{inputenc} 
\usepackage[T1]{fontenc}    
\usepackage{hyperref}       
\usepackage{url}            
\usepackage{booktabs}       
\usepackage{amsfonts}       
\usepackage{microtype}      
\usepackage{xcolor}         
\usepackage{amsmath}
\usepackage{graphicx}  
\usepackage{float}  
\usepackage{algorithm}
\usepackage{algorithmic}
\usepackage{multirow}
\usepackage{enumerate}
\usepackage{amsthm}
\usepackage{colortbl}

\usepackage{wrapfig}
\usepackage{multicol}
\usepackage{times}
\usepackage{soul}
\usepackage{url}
\usepackage{subcaption}

\journal{********}

\begin{document}

\begin{frontmatter}

\title{Towards One-shot Federated Learning: Advances, Challenges, and Future Directions}

\author[first]{Flora Amato}
\author[second]{Lingyu Qiu}
\author[third]{Mohammad Tanveer}
\author[second]{Salvatore Cuomo}
\author[second]{Fabio Giampaolo}
\author[second]{Francesco Piccialli}

\address[first]{
            Department of Electrical Engineering and Information Technology, University of Naples Federico II, Naples, Italy
            }

\address[second]{
               Department of Mathematics and Applications "R. Caccioppoli", University of Naples Federico II, Naples, Italy
            }
            
\address[third]{
Department of Mathematics, Indian Institute of Technology Indore, Simrol, Indore, Madhya Pradesh, India
            }

\begin{abstract}
One-shot FL enables collaborative training in a single round, eliminating the need for iterative communication, making it particularly suitable for use in resource-constrained and privacy-sensitive applications. This survey offers a thorough examination of One-shot FL, highlighting its distinct operational framework compared to traditional federated approaches. One-shot FL supports resource-limited devices by enabling single-round model aggregation while maintaining data locality. The survey systematically categorizes existing methodologies, emphasizing advancements in client model initialization, aggregation techniques, and strategies for managing heterogeneous data distributions. Furthermore, we analyze the limitations of current approaches, particularly in terms of scalability and generalization in non-IID settings. By analyzing cutting-edge techniques and outlining open challenges, this survey aspires to provide a comprehensive reference for researchers and practitioners aiming to design and implement One-shot FL systems, advancing the development and adoption of One-shot FL solutions in a real-world, resource-constrained scenario.
\end{abstract}




\end{frontmatter}

\setcounter{secnumdepth}{0} 

\section{1 Introduction}
\subsection{1.1 Background}

The development of deep neural networks (DNNs) has revolutionized the application of artificial intelligence across a wide range of both established and emerging fields, including healthcare~\cite{bhattamisra2023artificial,abdou2022literature}, finance~\cite{luo2024deep,xiang2022temporal}, and autonomous systems~\cite{shaheen2022continual,chen2021deep}. The evolution of deep neural networks (DNNs) has highlighted the need for decentralized approaches ensuring data privacy. Federated Learning (FL) addresses this need by enabling collaborative training of global models across distributed clients while ensuring that sensitive local data remains private. In FL, clients iteratively share and aggregate model parameters—rather than raw data—to preserve privacy while harnessing their collective knowledge of decentralized datasets. This approach aligns with growing demands for scalable, secure machine learning in applications ranging from healthcare to edge computing.
\subsubsection{1.1.1 Federated Learning Overview}
In federated learning, research focuses include communication efficiency, model aggregation methods, data heterogeneity, and privacy protection. As a privacy-preserving distributed machine learning paradigm, communication efficiency is one of the core bottlenecks of federated learning. Since FL relies on frequent communication between multiple clients and a central server to exchange model parameters, communication overhead often becomes the main limiting factor of system performance, especially in resource-constrained edge devices or bandwidth-limited network environments.
\subsubsection{1.1.2 Communication Costs in Federated Learning}
A traditional federated learning approach often requires a multi-round training process, where each client repeatedly communicates with a central server to exchange model updates. In a scenario involving $m$ clients and $n$ update rounds, this process results in the transmission of $O(mn)$ gradients or models, resulting in great potential risk for the leakage of local data and violation of the privacy-preserving principle of FL. The iterative nature of traditional federated learning introduces two critical challenges. First, repeated communication rounds heighten the risk of sensitive information exposure. Second, they incur significant communication overhead, which becomes particularly prohibitive in resource-constrained environments such as IoT networks or mobile devices~\cite{li2022datarobust}. These limitations hinder the scalability and practicality of federated learning in real-world applications, necessitating more efficient and privacy-preserving alternatives.

The original FL framework requires participants to engage in frequent and prolonged interactions with the central server, which can lead to high latency, energy consumption, and bandwidth usage. These challenges are exacerbated in scenarios where clients have limited computational resources or operate in environments with unreliable network connectivity. For instance, in healthcare applications, where data privacy is paramount~\cite{antunes2022federated}, or in industrial IoT systems~\cite{li2022effective}, where real-time decision-making is critical, such high communication costs may render traditional FL impractical.

To address these limitations, recent research has focused on minimizing communication overhead, while improving privacy guarantees without compromising model performance. Techniques such as model compression~\cite{shah2021model}, sparse updates~\cite{li2024fedsparse}, and federated distillation~\cite{li2019fedmd,lee2022preservation} have been developed to minimize data transmission during communication rounds. Among these, One-shot Federated Learning (One-shot FL)~\cite{guha2019one} has emerged as a promising approach. By completing the training process in a single communication round, One-shot FL significantly reduces both the risk of data leakage and communication costs. This single-round aggregation enhances system efficiency~\cite{allouah2024revisiting} while aligning closely with the privacy-preserving objectives of federated learning, making it a viable solution for real-world applications where efficiency and privacy are paramount.
\begin{figure}
    \centering
    \includegraphics[width=1\linewidth]{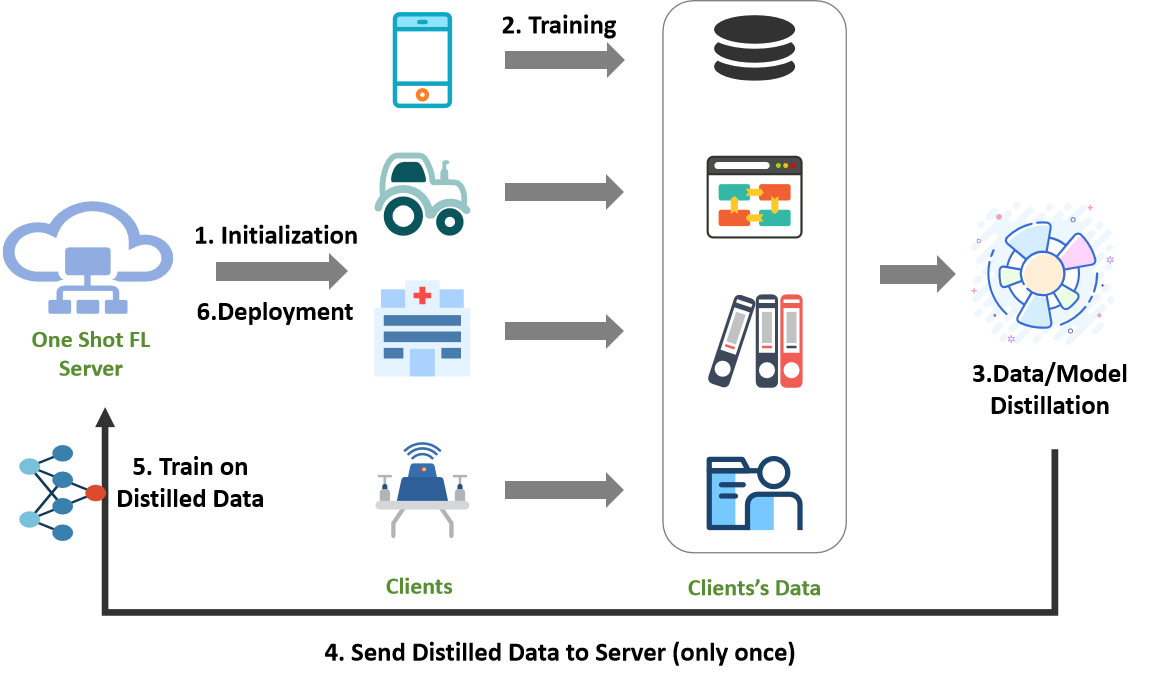}
    \caption{The training and deployment process of OSFL.}
    \label{fig:training}
\end{figure}

\subsubsection{1.1.3 One-shot Federated Learning Concept}
One-shot Federated Learning (One-shot FL) certainly differs from standard FL in some ways, but it remains firmly within the federated framework. One-shot FL limits communication to a single round while attempting to retain performance. However, performance often degrades under high statistical heterogeneity, and One-shot FL presents challenges in securing the entire training pipeline, particularly regarding protection against model poisoning attacks and ensuring the integrity of aggregated models. By synthesizing the relevant literature, this work provides a holistic view of the current state of the art in One-shot FL, while also identifying key areas for future research..
\subsection{1.2 Motivation}

The growing demand for scalable and efficient Federated Learning (FL) in distributed environments has highlighted communication overhead as a critical challenge.
This is particularly true in scenarios involving resource-constrained devices or latency-sensitive applications. Traditional FL methods rely on multiple communication rounds between clients and the central server, which can be both time-consuming and resource-intensive. In contrast, One-shot FL, which operates training within a single round of communication, has emerged as a promising approach to address these limitations. However, the unique requirements of One-shot FL, including model aggregation, optimization, and deployment feasibility, remain underexplored.

Existing surveys on FL primarily focus on model aggregation~\cite{qi2024model}, communication-efficient strategies~\cite{zhao2023towards}, or specific applications like IoT and edge devices~\cite{imteaj2021survey}. 
Few have thoroughly examined the unique intricacies of One-shot FL or provided a comprehensive framework to understand its capabilities, limitations, and future potential~\cite{liu2025one}. This gap highlights the need for a dedicated survey that systematically analyzes the progress, challenges, and opportunities within this emerging field.
\subsection{1.3 Contribution}

Given this reason, this survey provides a thorough investigation of One-shot Federated Learning, systematically addressing its methodologies, challenges, applications, and future directions. Our contributions are summarized as follows:

\textbf{Comprehensive Overview:} We examine the key principles and methodologies underpinning One-shot FL, focusing on its potential to reduce communication overhead while maintaining robust model performance.

\textbf{Current Progress}: By analyzing a wide range of recent studies, we summarize the state-of-the-art advancements in One-shot FL, including novel aggregation techniques, optimization strategies, and deployment scenarios.

\textbf{Practical Implications}: We discuss the feasibility of One-shot FL in real-world distributed environments, addressing requirements such as hardware capabilities, software frameworks, and system scalability.

\textbf{Future Directions}: Our survey identifies key challenges and proposes potential research directions to drive innovation in One-shot FL, paving the way for more efficient and scalable distributed learning systems.

By consolidating existing research and providing actionable insights, this survey serves as a valuable resource for researchers and practitioners, fostering further advancements and applications in the field of One-shot Federated Learning.

\subsection{1.4 Applicability}
\paragraph{1.4.1 Web3.0}
In the context of Web 3.0 applications, One-shot FL is a good solution to the transaction speed limitations of the contemporary commercial blockchain Ethereum (ETH)~\cite{wood2014ethereum}, and the high transaction costs (e.g., gas fees) on Web 3.0. One Shot FLW3~\cite{jiang2024ofl} introduces a new way to merge FL with Web 3.0 as data silos by actually participating in the collaborative machine learning process with incentives.
\paragraph{1.4.2 Autonomous Vehicles }
Federated Learning is particularly useful for object detection in {Autonomous Vehicles environments, where, for example, large amounts of data can be collected from various sources and used to train models that are robust to different weather conditions.
Selecting appropriate end devices, especially in autonomous vehicles that should participate in FL, is a major challenge to the adoption of FL. 
~\cite{10182876} proposes a real-time decision model based on trust-aware DRL to select trusted AVs, making the One-shot FL training process trustworthy.
\paragraph{1.4.3 IoT and Edge Computing}
~\cite{rjoub2022one} leverages model-free reinforcement learning and One-shot FL to create a smarter IoT device that can decide whether to automatically label samples or request true labels for a One-shot learning setting. This leads to a more efficient and effective solution for FL in IoT scenarios.

~\cite{song2023federated} achieves efficient federated learning by transferring the local distilled dataset to the server in a one-time manner, which can be used as a pre-trained model for personalization~\cite{song2023fedbevt}, fairness-aware learning~\cite{yu2020fairness}, etc.
\subsection{1.5 Survey structure}
\begin{figure*}
    \centering
    \includegraphics[width=1\linewidth]{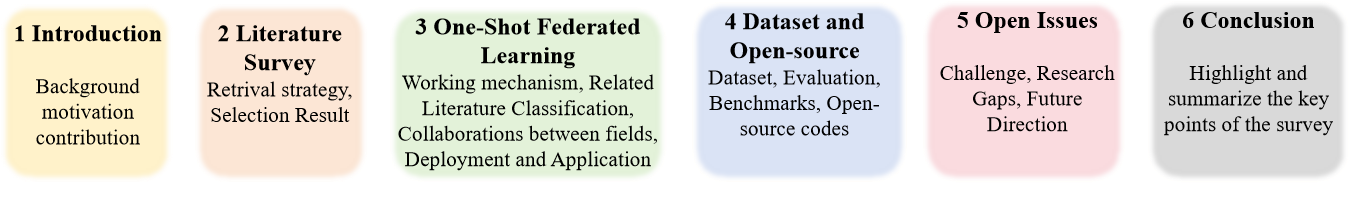}
    \caption{The structure of the survey}
    \label{fig:structure}
\end{figure*}

In this survey, the structure is organized as follows: 
Section 2 provides a detailed overview of the literature collection process, including the search terms, eligibility criteria, and evaluation methodology. 

Section 3 systematically examines the workflow of One-shot Federated Learning (One-shot FL), covering key aspects ranging from \textit{Optimization Algorithms} to \textit{Knowledge Distillation Based Methods} and \textit{Data Heterogeneity}.

Section 4 highlights open-source code and tools that facilitate the reproducibility of One-shot FL-related research. 

Section 5 addresses the current challenges in this domain and outlines potential directions for future research such as Scientific Machine Learning. 
\section{2 Related literature survey}
In this section, a thorough review of the existing literature was conducted, emphasizing studies in the domains of One-shot learning and federated learning (FL). The following sections outline the systematic methodology employed to identify relevant research articles and detail the rigorous selection process adopted to ensure the inclusion of high-quality, pertinent works.
\subsection{2.1 Retrieval strategy}
This survey rigorously follows the PRISMA guidelines to systematically outline the methods employed in this literature review. As the gold standard for systematic reviews and meta-analyses, PRISMA offers a robust framework that ensures the transparency, reliability, and reproducibility of our research. Specifically, we carefully defined the search terms, time frame, and scope to maintain consistency with the review's focus. We then screened the titles and abstracts to identify studies that met the eligibility criteria, followed by a comprehensive evaluation of the full texts for further validation. Ultimately, the inclusion of studies was determined based on their relevance to the review topic and their contribution of valuable insights or solutions to the research questions.
\subsubsection{2.1.1 Articles search}
This survey conducted a targeted search across three key databases: Scopus, arXiv, and OpenReview, which are used to collect current knowledge from a variety of sources. Scopus is a comprehensive database that provides a large number of peer-reviewed journal articles and conference papers in computer science and machine learning. Considering the novelty of our survey topic, this survey systematically uses arXiv and OpenReview to collect all relevant preprints and articles submitted to open review. arXiv provides a repository of preprints that have not yet been officially released. This approach of including both established databases and preprint sources ensures an exhaustive examination of the existing literature on our topic. Specifically, the search in the three databases uses a specific set of keywords: TITLE-ABS-KEY:
((‘‘One-shot learning’’ OR ‘‘Data-free’’ OR ‘‘One-shot communication’’) AND (‘‘federated learning’’)
AND NOT (‘‘few-shot’’ OR ‘‘zero-shot’’)), which consists of a combination of these terms (including their plural forms). To avoid increasing the workload of article screening, we excluded keywords related to few-shot and zero-shot because they are often mentioned in irrelevant literature.
\subsubsection{2.1.2 Eligibility criteria}
In alignment with the research topic and the PRISMA guidelines, we defined precise inclusion and exclusion criteria for the literature review process.

The following types of papers were excluded: (i) non-English language papers; (ii) duplicate studies; (iii) publications classified as "review articles" or "conference reviews."

The studies included in the review adhered to these standards: (i) at least one relevant search term appeared in the title, abstract, or keywords; (ii) through a detailed examination of both the abstract and full text, the paper was shown to be pertinent to edge deployment or One-shot learning; (iii) an in-depth assessment of the abstract and full text confirmed the paper's relevance to the application of single communication in Federated Learning (FL) or other distributed settings.
\subsubsection{2.1.3 Screening process}
\begin{figure}[htbp]
    \centering
    \includegraphics[width=1\linewidth]{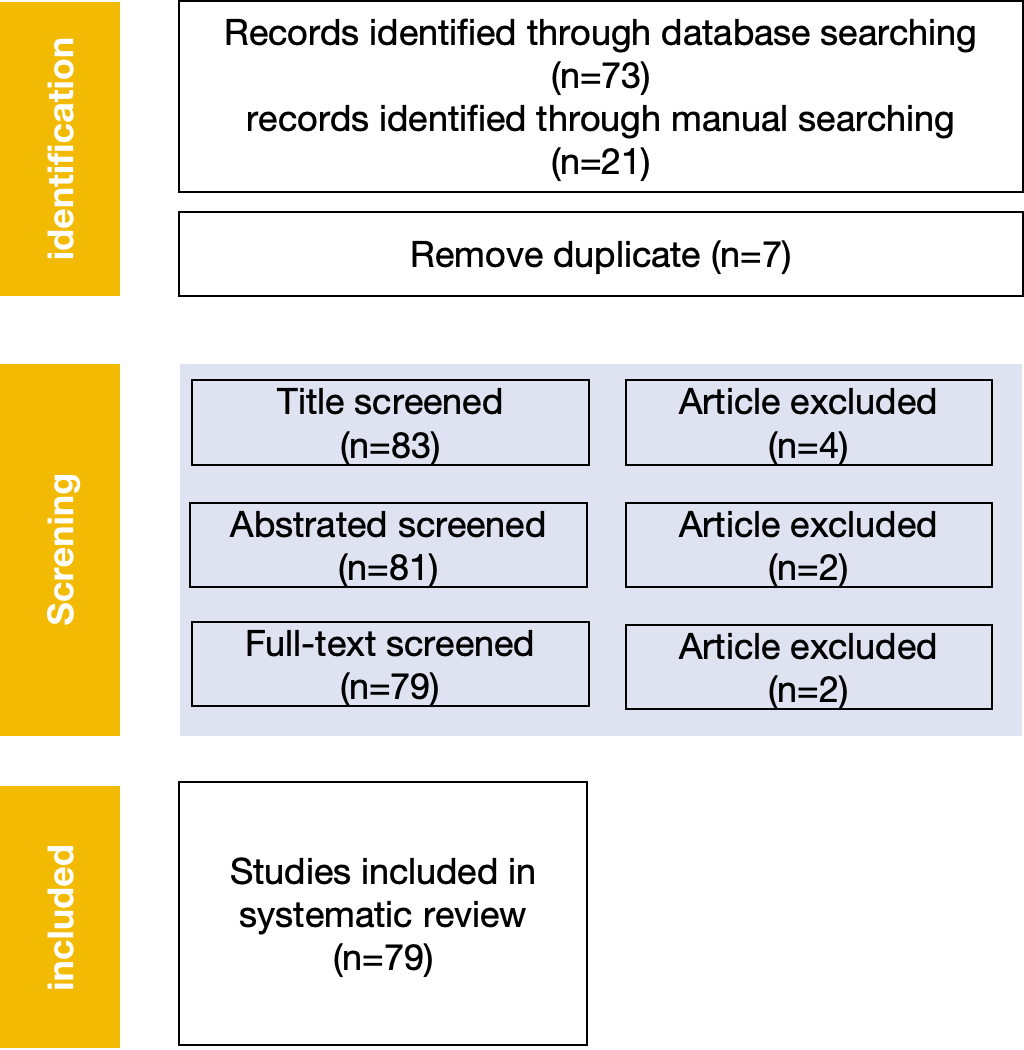}
    \caption{The methodology for retrieving research articles in this study follows a three-step process: identification, screening and eligibility, and inclusion. In the identification step, relevant keywords were selected to gather research articles related to the topic under investigation. During the screening and eligibility step, specific criteria were defined to filter out literature that did not align with the study's goals and objectives. Finally, in the inclusion step, the articles that satisfied the predefined criteria and matched the research requirements were incorporated into the study.}
    \label{fig:selection}
\end{figure}
To ensure the inclusion of high-quality and comprehensive
 studies, this survey followed a systematic approach to article selection. First, searches were conducted in Scopus and arXiv databases using carefully selected keywords. The search was then optimized by restricting the language to “English” and excluding document types that did not meet the predefined criteria, such as “conference review,” “comment,” “editorial,” and “letter.” A secondary screening step was performed using Python and excel tools to increase efficiency and speed. This involved running scripts to remove duplicate titles and filter out articles with “review” or “survey” in the title. In the final stage, independent reviewers among the authors thoroughly evaluated the abstracts and full texts against the established inclusion criteria. This multi-step process ensured a rigorous and comprehensive selection procedure. 

Figure \ref{fig:selection} outlines the number of papers at each stage. Initially, 73 papers were identified through the search query, and an additional 21 papers were found through manual search. After removing duplicates and review articles through Python, 79 papers were retained for further consideration. In the screening stage, multiple independent reviewers evaluated the titles, abstracts, and full texts, and 79 papers were finally selected.

\subsection{2.2 Selection results}
To provide a clear and systematic representation of the selection results, this survey employs a visualization approach that categorizes the data by publication year. This methodological choice enables a granular analysis of the temporal evolution and scholarly focus on One-shot Federated Learning (One-shot FL). By examining the distribution of publications over time and across prominent academic venues, we aim to uncover trends in research activity, identify key contributors, and assess the growing interest in this emerging paradigm. Such an analysis not only elucidates the developmental trajectory of One-shot FL but also highlights the academic communities and platforms that have played a pivotal role in shaping its discourse. This visualization serves as a foundational tool for understanding the field's progression and for guiding future research directions.

The results in Figure \ref{fig:number} demonstrate that One-shot Federated Learning (One-shot FL) has attracted significant scholarly attention since its introduction at ICML 2020, highlighting its critical importance and relevance. A temporal analysis shows a sharp increase in publications during 2023 and 2024, indicating a surge in research activity. This trend reflects the novelty of One-shot FL and confirms the growing interest in federated learning under strict communication constraints. The rise in publications further validates that communication-efficient FL, particularly with limited communication rounds, has become a prominent and rapidly advancing research area.
\begin{figure}[htbp]
    \centering
    \includegraphics[width=1\linewidth]{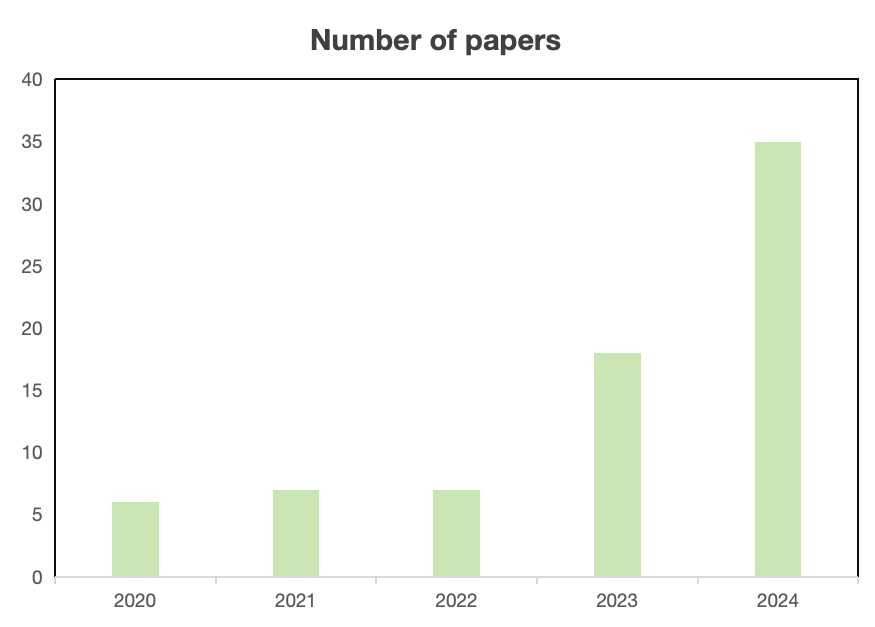}
    \caption{The number of included papers per year.}
    \label{fig:number}
\end{figure}

\section{3 One-shot Federated Learning}
\begin{figure*}[htbp]
    \centering
    \includegraphics[width=1\linewidth]{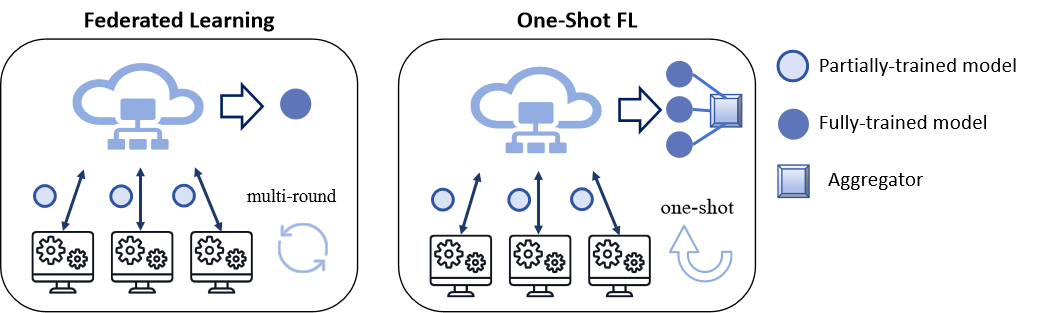}
    \caption{The pipeline of One-shot FL training.}
    \label{fig:pipeline}
\end{figure*}
In this section, we provide a comprehensive overview of One-Shot FL, covering its working mechanism, related literature classification, interdisciplinary collaborations, and deployment and applications.
\subsection{3.1 Working Mechanism}
The working mechanism of One-Shot FL is structured around three core components:(1) Problem Formulation, which defines the optimization objective for training a global model using decentralized data while minimizing communication rounds.
(2) Ensemble Aggregation, which combines knowledge from multiple client models into a unified global model.
(3) Synthetic Data Generation, which generates representative synthetic data to facilitate model training and aggregation without accessing raw client data.
In this session, this survey introduces the workflow of One-shot Federated Learning in detail.
\subsubsection{3.1.1 Problem Formulation}
Consider there have a set of clients $\mathcal{C}=\{c_1,c_2,...,c_n\}$, 
where $n = |\mathcal{C}|$ denotes the total number of clients. Each client $c_k \in \mathcal{C}$ possesses a local private dataset $\mathcal{D}^k = \{(\mathbf{x}_i, y_i)\}_{i=1}^{n_k}$, where $n_k = |\mathcal{D}^k|$ represents the number of local data samples $\mathbf{x}_i$ with the corresponding label $y_i$. The goal of One Shot Federated Learning(One-shot FL) is to train a global model with parameters $\boldsymbol{\theta}_S$ on the aggregated dataset $\mathcal{D} = \bigcup_{k=1}^n \mathcal{D}^k$ in a single communication round. This is formalized as:
\begin{equation}
    \min_{\boldsymbol{\theta}_S} \mathcal{L}(\boldsymbol{\theta}_S) \triangleq \frac{1}{|\mathcal{D}|} \sum_{(\mathbf{x}_i, y_i) \in \mathcal{D}} \ell_{\text{CE}}(f_S(\mathbf{x}_i; \boldsymbol{\theta}_S), y_i),
\end{equation}
where $\ell_{\text{CE}}(\cdot, \cdot)$ is the cross-entropy function, $f_S(\mathbf{x}_i; \boldsymbol{\theta}_S)$ is the server's prediction function, which outputs the logits (i.e., outputs of the last fully connected layer) of $\mathbf{x}_i$ given parameter $\boldsymbol{\theta}_S$.
\subsubsection{3.1.2 Ensemble Aggregation}
Noticeably, in One-shot Federated Learning, the original training sets $\mathcal{D}^k$ are inaccessible, and only pre-trained models parameterized by $\boldsymbol{\theta}_k$ are provided. To aggregate these models, we define an ensemble function:
\begin{equation}
    A_w(\mathbf{x}; \{\boldsymbol{\theta}_k\}_{k=1}^n) \triangleq \sum_{k=1}^n w_k f_k(\mathbf{x}; \boldsymbol{\theta}_k),
\end{equation}
where $f_k(\mathbf{x}; \boldsymbol{\theta}_k)$ is the prediction function of the $\boldsymbol{\theta}_k$-th client, outputting logits for input $\mathbf{x}$ 
given parameters $\boldsymbol{\theta}_k$. $\mathbf{w} = [w_1, w_2, \ldots, w_n]$  represents the weights assigned to each client's logits.

When $w_k = \frac{1}{n}$, the ensemble is the same as the averaged ensemble, while when $w_k = n_k / \sum_{k=1}^n n_k$, the ensemble becomes weighted according to the data amount. For simplicity, in the rest of the paper, we denote the ensemble as $A_w$ and its output logits for input $\mathbf{x}$ as $A_w(\mathbf{x})$; $\{\boldsymbol{\theta}_k\}_{k=1}^n$ means the output logits of the ensemble given $\mathbf{x}$.
\subsubsection{3.1.3 Synthetic Data Generation}
To aggregate pre-trained models $\{\boldsymbol{\theta}_k\}_{k=1}^n$ into one server model $\boldsymbol{\theta}_S$, existing works mostly follow a two-stage framework. The first is to synthesize data $\mathcal{D}_S$ based on the ensemble output. In particular, giving a random noise $\mathbf{z}$ sampled from a standard Gaussian distribution and a random uniformly sampled label $y_S$, the generator $G(\cdot)$ with $\boldsymbol{\theta}_G$ is responsible for generating the data $\mathbf{x}_S = G(\mathbf{z})$, forming the synthetic dataset $\mathcal{D}_S$. Typically, to make sure the synthetic data can be classified correctly with a high probability by the ensemble $A_w$, the following loss is adopted:
\begin{equation}
    \mathcal{L}(\boldsymbol{\theta}_G) \triangleq \frac{1}{|\mathcal{D}_S|} \sum_{(\mathbf{x}_S, y_S) \in \mathcal{D}_S} \ell_{\text{CE}}(A_w(\mathbf{x}_S), y_S). \label{eq:3}
\end{equation}

After getting the synthetic dataset $\mathcal{D}_S$ based on the generator in Eq.\ref{eq:3}, One Shot FL intends to distill the ensemble $A_w$ into the final server model $\boldsymbol{\theta}_S$ with the help of these synthetic data, as in:
\begin{equation}
    \min_{\boldsymbol{\theta}_S} \mathcal{L}(\boldsymbol{\theta}_S) \triangleq \frac{1}{|\mathcal{D}_S|} \sum_{(\mathbf{x}_S, y_S) \in \mathcal{D}_S} \ell_{\text{KL}}(A_w(\mathbf{x}_S), f_S(\mathbf{x}_S; \boldsymbol{\theta}_S))
    \label{eq:4}
\end{equation}
where $\ell_{\text{KL}}(\cdot, \cdot)$ denotes the Kullback-Leibler (KL) divergence.

Existing works illustrate that the performance of the server model is intrinsically related to the synthetic data $\mathcal{D}_S$ and the ensemble $A_w$, which can also be concluded according to Eq.\ref{eq:4}
\subsection{3.2 Related Literature Classification}
To systematically organize and evaluate the growing body of research on One-Shot Federated Learning (One-Shot FL), we propose a unified taxonomy that categorizes existing work into four key areas: Theory-Based Optimization Algorithms, Knowledge Distillation-Based Ensemble Methods, Data Heterogeneity, and Adversarial Robustness. This classification framework provides a structured overview of the field, highlighting both progress and gaps in current research.
\subsubsection{3.2.1 Theory-based Optimization Algorithms}
FedLPA~\cite{liufedlpa} trains global models using block-diagonal empirical Fisher information matrices without requiring extra auxiliary datasets or exposing any private label information, which captures the statistics of the posteriors of each layer effectively.
qFedAvg~\cite{zhao2023non} explores the non-IID problem in quantum federated learning with theoretical and numerical analysis. It further proves that the global quantum channel can be accurately decomposed into local channels trained by each client with the help of local density estimation. Thus, a quantum federated learning framework on non-IID data with one-time communication complexity is proposed.
FedFisher~\cite{jhunjhunwala2024fedfisher} analyzes global model loss for overparameterized ReLU networks. It makes use of Fisher information matrices computed on local client models and reformulates it as a posterior inference problem then provides theoretical guarantees for One-shot Federated Learning through theoretical analysis of the case of over-parameterized two-layer neural networks.
FuseFL~\cite{tang2024fusefl} using causality to analyze the gap between multi-round FL and One Shot FL and alleviate data heterogeneity of One Shot FL.
OSFL-DQN~\cite{rjoub2022one} combines model-free reinforcement learning with One-shot FL to design a smarter way to integrate model-free deep reinforcement learning technology into IoT devices, where local training is performed on IoT devices and global aggregation is done at the edge server level.
Task Arithmetic~\cite{tao2024taskarithmeticlensoneshot} studies the task algorithms of multi-task learning as a one-time federated learning problem, providing a new theoretical perspective for task algorithms and an improved practical method for model merging.
FedELMY~\cite{wang2024one} leverage the potential of local model diversity, introduce a local model pool for each client that comprises diverse models generated during local training, and propose two distance measurements to further enhance the model diversity and mitigate the effect of non-IID data. 
\subsubsection{3.2.2 Knowledge Disllation based Ensemble Method}
There are many studies that apply knowledge distillation techniques to one-time federated learning, especially model ensemble:
DENSE~\cite{zhang2022dense} proposes a data-free approach in distillation, which utilizes a generator to create synthetic datasets on the server side, which first trains the discriminative classifiers on the clients to convergence and iterative optimization between training a GAN-based network to for generating synthetic data and using the synthetic data to distill the ensemble of client.
FEDCVAE~\cite{heinbaugh2023data} mainly addresses the limitations of One-shot FL methods, such as often degrading performance under high statistical heterogeneity, failing to promote pipeline security, or requiring auxiliary public datasets. The authors propose two variants based on FEDCVAE: FEDCVAE-ENS and its extension FEDCVAEKD. What these two methods have in common is that they both reformulate the local learning task using conditional variational autoencoders (CVAE) to address high statistical heterogeneity. This is because CVAE can easily learn simplified data distributions, so both methods train CVAE locally to capture the narrow conditional data distribution that arises in high statistical heterogeneity settings.
The respective contributions of the two variants are that the decoder of FEDCVAE-ENS is integrated, while FEDCVAE-KD is compactly aggregated, i.e., knowledge distillation is used to compress the client decoder set into a single decoder~\cite{heinbaugh2023data}. Both methods use the method of moving the center of the CVAE prior distribution to safely combine heterogeneous local models safely.

FedOV~\cite{diao2023towards} mainly addresses the case of comprehensive label skew, marking another step forward in the evolution of One-shot FL methods.
FedDISC~\cite{yang2024exploring} relies on auxiliary pre-trained model CLIP, however their reliance may not always align with its practicality or applicability for different scenarios.
MAEcho~\cite{su2023one} emphasizes the standardized addition between layer width parameters in the local model aggregation process, and shares the orthogonal projection matrices of client features with the server to optimize the global model parameters.
Dataset distillation anonymously maps the distilled dataset from the original client data~\cite{wang2020federated,dong2022privacy}, thus maintaining the privacy advantage of federated learning without any exposure. From the data ensemble technique, 
FedKT~\cite{li2020practical} achieves differential privacy guarantees by utilizing the knowledge transfer technique to improve the ensemble on the public data.
DOSFL~\cite{Zhou_Pu_Ma_Li_Wang_2020} is proposed to transmit the distilled local dataset for the server instead of using public data.
FedD3~\cite{song2023federated} integrates dataset distillation instances which provides the additional benefit to be able to balance the trade-off between accuracy and communication cost.
Co-Boosting~\cite{dai2024enhancing} improves the ensemble both data and model side when doing the distillation to improve the performance. 
FedDiff~\cite{mendieta2024navigating} exploring diffusion models in One-shot Federated Learning and improves generated sample quality under DP settings. 
\subsubsection{3.2.3 Data Heterogeneity}
The non-IID data problem arises from not taking into account the global data distribution.
XorMixFL~\cite{shin2020xor} proposes a privacy-preserving XOR-based mixup data augmentation technique. It maintains data privacy by distorting the original samples using bitwise XOR operations in both the encoding and decoding processes. Specifically, it involves collecting encoded data samples from other devices and decoding them using only data from each device. The decoding provides synthetic but realistic samples before inducing an IID dataset for model training.
However, these approaches still do not fully consider data heterogeneity~\cite{li2020practical,shin2020xor}, while some approaches face difficulties under high data heterogeneity~\cite{Zhou_Pu_Ma_Li_Wang_2020}.
FedOpt~\cite{dupuy2022learnings} studies popular FL algorithms on non-I.I.D. data and explores several sampling strategies and it was shown that non-uniform sampling is better than standard uniform sampling.
PersAvg~\cite{garin2023personalized} proposes to solve the problem of unbalanced training sample size and heterogeneous distribution among nodes through trade-off between the use of local and other nodes's information.
FedD3~\cite{song2023federated} incorporating dataset distillation by transferring the local distilled dataset to the server in a one-time manner achieves much better results due to the broader data resource.
FEN~\cite{allouah2024revisiting} is a hybrid of One Shot FL and standard FL using a two-stage aggregator model, which supports model heterogeneity and enables rapid client unlearning. However, it faces the memory limitation required to store the aggregator-trained ensemble models on the client device.
FedGM~\cite{chen2024one} devise a method to extract knowledge from each client's data by creating a synthesized dataset while applying label differential privacy.
\subsubsection{3.2.4 Adversarial Robustness}
The decentralized nature of federated learning also makes it vulnerable to backdoor attacks, where malicious actors can embed hidden vulnerabilities in the model. It is critical to effectively address these threats, especially given the impracticality of iterative and resource-intensive detection methods in federated learning environments.
Danilenka~\cite{danilenka2023one} research One-shot Federated Learning in the context of aggregation of the clients' updated models and the possibility of using adversarial images as a source of client-picking guidance and performance improvement in the presence of label skew non-IID data.
~\cite{pan2024one} proposes a new framework for one-time backdoor removal that combines anomaly detection techniques with model update aggregation strategies, without the need for extensive data access or communication between participants.
~\cite{andrew2023one} provides a provably correct estimate of privacy loss under the Gaussian mechanism. This approach eliminates the shortcomings of these One-shot Federated Learning techniques that are difficult to deploy at scale in practice, especially in federated settings where model training may take days or weeks.~\cite{andrew2023one} suggesting that adding a modest amount of noise is sufficient to prevent leakage.
In order to comprehensively compare the One-shot FL methods, this survey evaluates the communication performance of the methods from multiple aspects.
\begin{itemize}
    \item Data: (Encoded data \& no label distributions or statistics Sent)
    \item Model: (High model performance on both IID and NO-IID Data)
    \item Communication: Low Communication Cost
    \item Affiliate Resources (Public dataset \& Auxiliary Models Used)
\end{itemize}
\begin{table*}[htbp]
    \centering
    \small
    \begin{tabular}{c|ccccccc}
    \toprule
       Paper  & Data&Label&Performance& Computation&Public dataset&Auxiliary Models\\
       \midrule
      FedKT~\cite{li2020practical}   & \checkmark& \checkmark&& \checkmark&& \checkmark\\

      FedOV~\cite{diao2023towards}  & \checkmark & \checkmark& & \checkmark & \checkmark & \checkmark\\

      FedDISC~\cite{yang2024exploring}  & \checkmark& & \checkmark & \checkmark & \checkmark& \\

      XorMixFL~\cite{shin2020xor}& & \checkmark & \checkmark & \checkmark& & \checkmark\\

      DOSFL~\cite{Zhou_Pu_Ma_Li_Wang_2020}& & \checkmark  & & \checkmark & \checkmark & \checkmark\\

      DENSE~\cite{zhang2022dense} & \checkmark & \checkmark & \checkmark& & \checkmark & \checkmark  \\

      FedCADO~\cite{Yang_Su_Li_Xue_2023} & \checkmark & \checkmark & \checkmark & \checkmark & \checkmark\\

      FedCVAE~\cite{heinbaugh2023data} & \checkmark& & \checkmark & \checkmark & \checkmark & \checkmark\\
      
      MAEcho~\cite{su2023one} & \checkmark & \checkmark & \checkmark & \checkmark & \checkmark & \checkmark\\

      FedLPA~\cite{liufedlpa} & \checkmark & \checkmark & \checkmark & \checkmark & \checkmark & \checkmark\\
      
      FedFisher~\cite{jhunjhunwala2024fedfisher} & \checkmark & \checkmark & \checkmark & \checkmark & \checkmark & \checkmark\\
    \bottomrule
    \end{tabular}
    \caption{Comparsion the research papers in data, computation, model and affiliate resource persepactive}
    \label{tab:my_label}
\end{table*}
\subsection{3.3 Collaborations between fields}
The development of One-shot Federated Learning (One-shot FL) relies heavily on interdisciplinary collaboration, integrating expertise from cryptography, machine learning, and distributed systems to tackle its distinct challenges. Cryptography ensures secure aggregation and privacy-preserving mechanisms, safeguarding sensitive data during the single communication round. Machine learning introduces advanced optimization techniques and model architectures designed for One-shot FL's constraints, enabling efficient learning from decentralized, heterogeneous datasets. Distributed systems provide scalable frameworks and resource management strategies to address the complexities of large-scale deployments. By combining insights from these fields, One-shot FL can address critical limitations, such as communication efficiency, data heterogeneity, and privacy-accuracy trade-offs, unlocking robust applications in healthcare, IoT, and edge computing. This subsection examines how such interdisciplinary efforts drive progress in One-shot FL, emphasizing key synergies and future research directions.

Cryptography plays a critical role in One-shot Federated Learning, enabling secure communication and robust protection of sensitive data between clients and the server.
Federated Learning can be augmented to satisfy user-level differential privacy (DP)~\cite{Dwork_Roth_2013}, ensuring rigorous privacy guarantees.
Privacy estimation techniques for differentially private (DP) algorithms can be used to compare against analysis bounds or to empirically measure privacy loss in settings where known analysis bounds are not strict. Recent work by~\cite{karthikeyan2024opa} addresses the practical challenges of DP in model training by introducing privacy estimation techniques. These techniques enable efficient auditing or estimation of privacy loss during the training process, eliminating the need for prior knowledge of the model architecture, task, or DP algorithm. The method demonstrates provably correct privacy loss estimates in Gaussian regimes and robust performance under adversarial threats.
To address the challenge of large-scale deployment of privacy-preserving techniques in practice,~\cite{andrew2023one}proposes a privacy-preserving method that does not require any prior knowledge about model architecture, tasks, or differentially private training algorithms.
Their results show that in common scenarios—where only the final model is released—moderate noise injection suffices to mitigate data leakage risks~\cite{andrew2023one}. This finding simplifies the practical implementation of privacy-preserving One-Shot FL in resource-constrained environments.

\subsection{3.4 Deployment and Application}
The deployment and application of One-shot Federated Learning (One-shot FL) has gained significant attention due to its ability to address communication efficiency, privacy protection, and scalability issues in distributed learning scenarios. In this section, we explore the practical deployment strategies and real-world applications of One-shot Federated Learning, highlighting its transformative potential in various fields.
\subsubsection{3.4.1 Healthcare}
One-Shot Federated Learning (One-Shot FL) enables collaborative model training in privacy-sensitive domains like medical diagnosis and drug discovery while minimizing communication overhead.  In healthcare, hospitals use One-Shot FL to train diagnostic models, such as COVID-19 detection from chest X-rays, without sharing patient data, ensuring compliance with regulations like HIPAA~\cite{shafik2024digital}.
Similarly, pharmaceutical companies leverage low communication costs to predict molecular properties or optimize drug candidates, reducing trial expenses while protecting proprietary data. These applications highlight One-Shot FL's ability to balance privacy, efficiency, and scalability in distributed learning~\cite{kang2023one}.
\subsubsection{3.4.2 Recommentation System}
One-FedCF~\cite{eren2022one} first implements One-shot Federated Learning for CF and recommenders.One-FedCF~\cite{eren2022one} obtains client-specific recommendations with only one-pair communication between the server and its clients after a small amount of initial communication. One-FedCF mainly exploits joint non-negative matrix factorization and information retrieval transfer using latent representation to represent the utility of CF in a group setting for recommendation~\cite{o2001polylens}.
\subsubsection{3.4.3 Satellite Communications}
At altitudes of 160-2,000 km above the Earth’s surface, LEO satellites are typically equipped with sensors and high-resolution cameras to collect a large amount of mobility-related data, such as tracking of hurricanes, weather forecasts~\cite{Martin_Park_Camps_2021} and forest fire movements~\cite{Barmpoutis_Papaioannou_Dimitropoulos_Grammalidis_2020} and monitoring of cloud movement, flood conditions, migration of large animals across geographic areas and aircraft tracking. 
LEOShot~\cite{elmahallawy2023one} is a novel One-shot Federated Learning approach for LEO satellite constellations, which can directly applied to SatCom.
\begin{figure}
    \centering
    \includegraphics[width=1\linewidth]{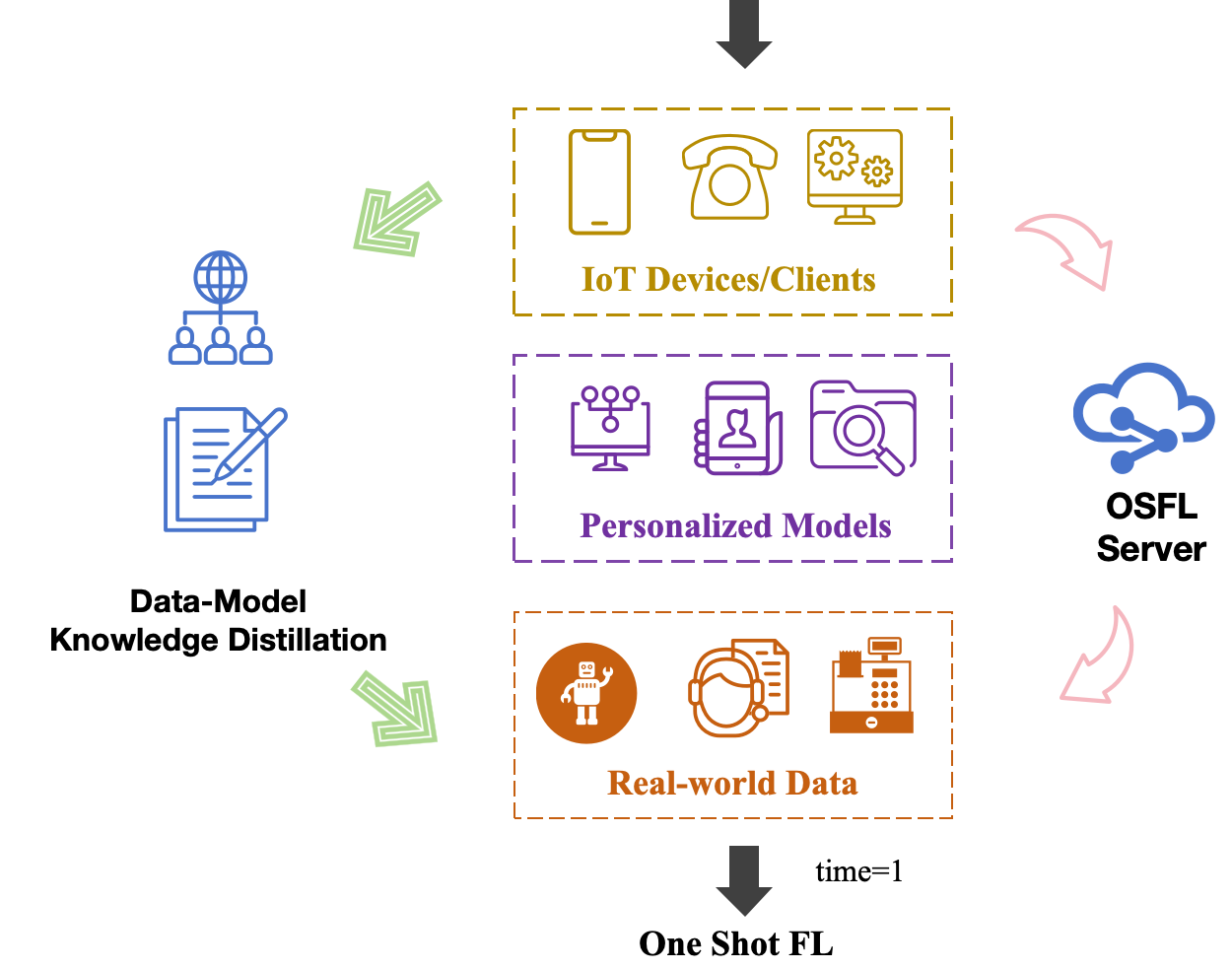}
    \caption{Edge Deployment and Application Scenarios of One-shot Federated Learning in Real-World Systems}
    \label{fig:deployment}
\end{figure}
\section{5 Dataset and open-source codes}
In this section, we provide a comprehensive overview of the datasets and open-source codes referenced in the surveyed papers. These datasets span diverse scenarios and applications, offering valuable opportunities to advance One-shot Federated Learning. Moreover, the availability of open-source codes enhances research reproducibility and establishes robust benchmarks to guide future advancements in the field.
\subsection{5.1 Datasets}
This section mainly introduces the experimental settings and commonly used datasets of One Shot Federated Learning. In Table \ref{tab:dataset}, we introduce the basic information of the datasets commonly used in the One Shot Federated Learning algorithm experiment, including the MNIST dataset~\cite{lecun1998gradient} containing binary images of handwritten digits, and the CIFAR10 dataset~\cite{krizhevsky2009learning} containing 60,000 color images of 10 classes, 6,000 images per class. The CIFAR100 dataset~\cite{krizhevsky2009learning} is similar to the CIFAR10 dataset, except that CIFAR100~\cite{krizhevsky2009learning} has 100 classes, 600 images per class, 500 training images, and 100 test images per class. Tiny-ImageNet~\cite{krizhevsky2009learning} contains 100,000 images of 200 classes, 500 training images, 50 validation images, and 50 test images per class. For the size of the image, the input dimensions of MNIST~\cite{lecun1998gradient}, FMNIST~\cite{xiao2017fashion}, SVHN~\cite{netzer2011reading}, and CIFAR-10~\cite{krizhevsky2009learning} are 784, 784, 3,072, and 3,072, respectively.
SODA10M~\cite{han2021soda10m} is the large-scale 2D Self/semisupervised Object Detection Dataset for Autonomous driving. The SODA10M~\cite{han2021soda10m} contains 10 million unlabeled images and 20000 labeled images with 6 sample object categories. The images are taken every 10s per frame over the course of 27833 driving hours in various weather conditions, times, and location scenarios in 32 distinct cities to increase diversity.

\textbf{Real-world Datasets}:
There are several researchers~\cite{garin2023personalized,zhang2022dense,dupuy2022learnings} have used datasets in special fields such as healthcare. In this survey, we summarize some commonly used datasets for introduction.


CT-Scan~\cite{graf20112d} is composed of CT-Scan images coming from different patients. Each image corresponds to a relative position in the patient's body (between 0 and 100). The task is to predict this relative position from the image. 
GreenHouseGas~\cite{lucas2015designing} (GHG) contains time series data from different geographic regions in California. There are 15 GHG time series and one synthetic GHG time series for each geographic region. The model objective is to predict the last time series from the first 15 time series.
Space-ga~\cite{domingos2000unified} reports voting results for the 1980 U.S. presidential election. The task is to predict the voter turnout in a county based on several socioeconomic characteristics such as population size and average income.

\begin{table*}[htbp]
    \centering
    \begin{tabular}{c|cccccc}
    \toprule
               Dataset &opensource\\
        \midrule
        Fashion-MNIST~\cite{xiao2017fashion} &\href{https://huggingface.co/datasets/zalando-datasets/fashion_mnist}{https://huggingface.co/datasets/zalando-datasets/fashion\_mnist} \\
       MNIST~\cite{lecun1998gradient}  & \href{https://paperswithcode.com/dataset/mnist}{https://paperswithcode.com/dataset/mnist}\\
       CIFAR10~\cite{krizhevsky2009learning}& \href{https://huggingface.co/datasets/uoft-cs/cifar10}{https://huggingface.co/datasets/uoft-cs/cifar10} \\
       CIFAR100~\cite{krizhevsky2009learning}& \href{https://huggingface.co/datasets/uoft-cs/cifar100}{https://huggingface.co/datasets/uoft-cs/cifar100}\\
       Tiny-ImageNet~\cite{krizhevsky2009learning}&\href{https://huggingface.co/datasets/zh-plus/tiny-imagenet}{https://huggingface.co/datasets/zh-plus/tiny-imagenet} \\
       SVHN~\cite{netzer2011reading}&\href{https://huggingface.co/datasets/ufldl-stanford/svhn}{https://huggingface.co/datasets/ufldl-stanford/svhn} \\
       CINIC-10~\cite{darlow2018cinic}&\href{https://huggingface.co/datasets/flwrlabs/cinic10}{https://huggingface.co/datasets/flwrlabs/cinic10} \\
       FOOD101~\cite{bossard2014food}&\href{https://huggingface.co/datasets/ethz/food101}{https://huggingface.co/datasets/ethz/food101}\\

       SODA10M~\cite{han2021soda10m}&\href{https://soda-2d.github.io/}{https://soda-2d.github.io/}\\
       \midrule
       Energy~\cite{electricityloaddiagrams20112014_321}&\href{https://archive.ics.uci.edu/ml/datasets/ElectricityLoadDiagrams20112014}{https://archive.ics.uci.edu/ml/datasets/ElectricityLoadDiagrams20112014}\\

       GreenHouseGas~\cite{tubiello2013faostat}&\href{https://archive.ics.uci.edu/dataset/328/greenhouse+gas+observing+network}{https://archive.ics.uci.edu/dataset/328/greenhouse+gas+observing+network}\\

       CT-Scan~\cite{graf20112d} &\href{https://archive.ics.uci.edu/ml/datasets/Relative+location+of+CT+slices+ on+axial+axis}{https://archive.ics.uci.edu/ml/datasets/Relative+location+of+CT}\\

       Space-ga~\cite{domingos2000unified}&\href{https://mljar.com/machine-learning/use-case/space-ga/}{https://mljar.com/machine-learning/use-case/space-ga/}\\
       \bottomrule
    \end{tabular}
    \caption{The datasets commonly used in One-shot FL research experiments. Above the horizontal line are common image classification datasets, and below the horizontal line are special datasets based on different scene settings.}
    \label{tab:dataset}
\end{table*}
\subsection{5.2 Efficiency Evaluation}
In One-shot Federated Learning (One-shot FL), optimizing performance and efficiency is crucial, particularly when deploying on resource-constrained devices like IoT devices and mobile phones. One-shot Federated Learning aims to minimize communication rounds to a single round, which introduces unique challenges and opportunities for optimization. With theses predefined evaluation metrics can formalize and analyze which One-shot Federated Learning approach is best suited for this specific telecommunication use case

The performance of One-shot Federated Learning models is commonly assessed using several key metrics and experiment settings:
\begin{itemize}
    \item (1) Data Heterogeneity: Data heterogeneity results with various $Dir(\alpha)$ partitions.

\item (2) Number of Clients: observe the effect of increasing the distributed nature of the data across the client network.

\item (3) Resource Requirements: including FLOPs and parameter count, for each method deployed on a single client even when deployed on hardware with modest computational capabilities.
\end{itemize}

\subsection{5.3 Benchmarks}

A comprehensive benchmark plays a pivotal role in evaluating the performance, robustness, and scalability of algorithms under realistic conditions. In this section, we include the benchmark often used in One-shot FL experiments.

NIID-Bench\footnote{https://github.com/Xtra-Computing/NIID-Bench}~\cite{li2022federatedbench} is a benchmark for federated learning algorithms under non-IID data distribution scenarios. NIID-Bench implement 4 federated learning algorithms including FedAvg~\cite{mcmahan2017communication}, FedProx~\cite{li2020federated2020}, SCAFFOLD~\cite{karimireddy2019scaffold} \& FedNova~\cite{wang2020tackling}, 3 types of non-IID settings such as label distribution skew, feature distribution skew \& quantity skew, 9 datasets (MNIST~\cite{lecun1998gradient} , CIFAR100~\cite{krizhevsky2009learning}, Fashion-MNIST~\cite{xiao2017fashion}, SVHN~\cite{netzer2011reading}, Generated 3D dataset, FEMNIST~\cite{caldas2018leaf}, (adult, rcv1, covtype)\footnote{https://www.csie.ntu.edu.tw/~cjlin/libsvmtools/datasets/}.

FLamby\footnote{https://github.com/owkin/FLamby/}~\cite{ogier2022flamby} is a cross-silo
federated learning benchmark in realistic healthcare, which contains 7 datasets: Fed-Camelyon16~\cite{litjens20181399}, 
Fed-LIDC-IDRI~\cite{armato2011lung,clark2013cancer,armato2015data},
Fed-ISIC2019~\cite{codella2018skin,combalia2019bcn20000},
Fed-TCGA-BRCA~\cite{federated2022tensorflow},
Fed-Heart-Disease~\cite{perez2021torchio},
Fed-IXITiny~\cite{perez2021torchio},
Fed-KITS2019~\cite{heller2021state,heller2019kits19},
which aims to give more visibility to such isolated cross-silo initiatives has been widely used in One-shot FL research~\cite{allouah2024revisiting}.


\subsection{5.4 Open-source codes}

To facilitate the effective deployment of One-shot Federated Learning (One-shot FL),  we have curated a comprehensive list of open-source projects referenced in the studies surveyed. These resources not only enable the replication of existing research but also provide crucial references for steering future investigations within this domain.

\begin{table*}[h]
    \centering
    \begin{tabular}{ccc}
    \toprule
      Paper   & Year & Link \\
      \midrule
     ~\cite{wang2023data}   & 2023&\href{https://github.com/NaiboWang/Data-Free-Ensemble-Selection-For-One-shot-Federated-Learning}{https://github.com/NaiboWang/Data-Free-Ensemble-Selection-For-OSFL}\\
      
     ~\cite{zhu2021data}&2021&\href{https://github.com/zhuangdizhu/FedGen}{https://github.com/zhuangdizhu/FedGen}\\
     ~\cite{luo2024dfdg}&2024&
      \href{https://anonymous.4open.science/r/DFDG-7BDB}{https://anonymous.4open.science/r/DFDG-7BDB}\\
     ~\cite{jhunjhunwala2024fedfisher}&2024&\href{https://github.com/Divyansh03/FedFisher}{https://github.com/Divyansh03/FedFisher}\\
      
~\cite{liufedlpa}&2024&\href{https://openreview.net/attachment?id=I3IuclVLFZ&name=supplementary_material}{https://openreview.net/attachment?id=I3IuclVLFZ\&name=supplementary\_material}\\

~\cite{tang2024fusefl}&2024&
    \href{https://github.com/wizard1203/FuseFL}{https://github.com/wizard1203/FuseFL}\\

   ~\cite{dennis2021heterogeneity}&2021&
    \href{https://github.com/metastableB/kfed/}{https://github.com/metastableB/kfed/}\\

   ~\cite{garin2023incidence}&2023&
    \href{https://www.ipol.im/pub/art/2023/440/}{https://www.ipol.im/pub/art/2023/440/}\\

   ~\cite{humbert2024marginal}&2024&
    \href{https://github.com/pierreHmbt/One-shot-FCP}{https://github.com/pierreHmbt/One-shot-FCP}\\

   ~\cite{mendieta2024navigating}&2024&
    \href{https://github.com/mmendiet/FedDiff}{https://github.com/mmendiet/FedDiff}\\

   ~\cite{zhao2023non}&2023&
    \href{https://github.com/haimengzhao/quantum-fed-infer}{https://github.com/haimengzhao/quantum-fed-infer}\\

   ~\cite{zhang2024one}&2024&\href{https://github.com/Carkham/FedSD2C}{https://github.com/Carkham/FedSD2C}\\

   ~\cite{chen2024one}&2024&\href{https://github.com/baldcodeman/One-shot-Federated-Learning-with-Label-Differential-Privacy}{https://github.com/baldcodeman/One-shot-Federated-Learning-with-Label-DP}\\

   ~\cite{su2023one}&2023&\href{https://github.com/FudanVI/MAEcho}{https://github.com/FudanVI/MAEcho}\\
~\cite{salehkaleybar2021one}&2021&\href{https://github.com/sabersalehk/MRE_C}{https://github.com/sabersalehk/MRE\_C}\\

   ~\cite{wang2024one}&2024&\href{https://github.com/NaiboWang/FedELMY}{https://github.com/NaiboWang/FedELMY}\\

   ~\cite{zeng2024one}&2024&\href{https://github.com/zenghui9977/IntactOne Shot FL}{https://github.com/zenghui9977/IntactOne Shot FL}\\

   ~\cite{li2020practical}&2020&\href{https://github.com/QinbinLi/FedKT}{https://github.com/QinbinLi/FedKT}\\
    
~\cite{allouah2024revisiting}&2024&
    \href{https://github.com/sacs-epfl/fens}{https://github.com/sacs-epfl/fens}\\

~\cite{diao2023towards}&2023&
    \href{https://github.com/Xtra-Computing/FedOV}{https://github.com/Xtra-Computing/FedOV}\\
~\cite{yurochkin2019bayesian}&2019&
    \href{https://github.com/IBM/probabilistic-federated-neural-matching}{https://github.com/IBM/probabilistic-federated-neural-matching}\\

~\cite{andrew2023one}&2024&
    \href{https://github.com/google-research/federated/tree/master/one_shot_epe}{https://github.com/google-research/federated/tree/master/one\_shot\_epe}\\
    \bottomrule
    \end{tabular}
    \caption{Summary of One-shot Federated Learning Algorithms (Publication Year and Open Source Availability)}
    \label{tab:github_osfl}
\end{table*}

\section{6 Open issue}\label{sec:Open}

\begin{figure}
    \centering
    \includegraphics[width=\linewidth]{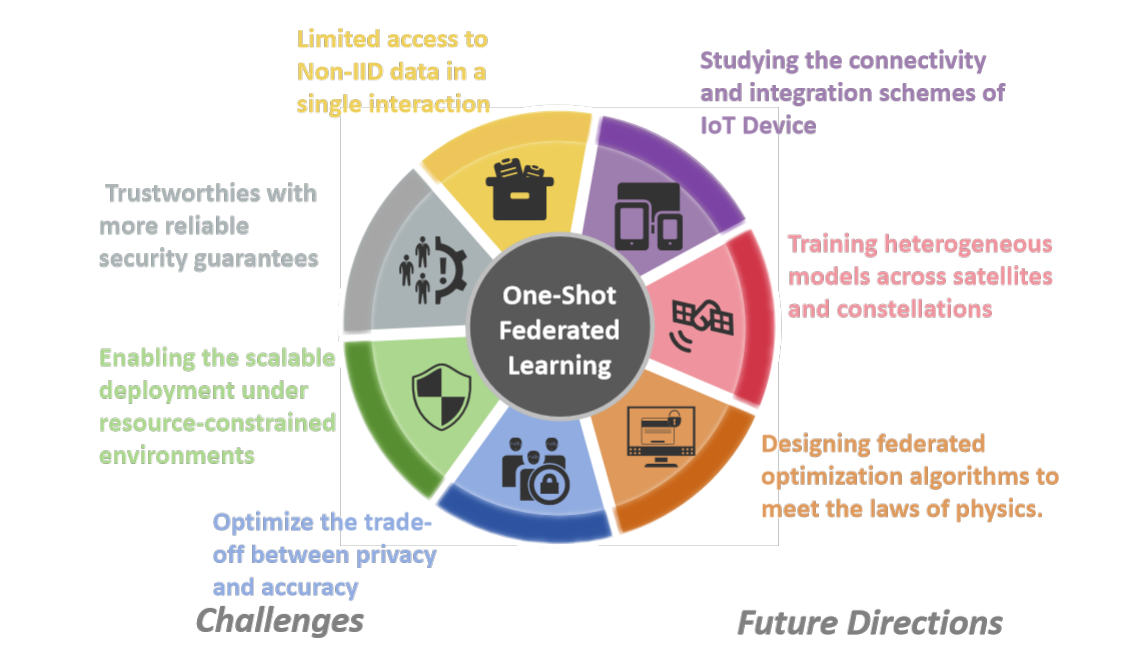}
    \caption{Challenges and future directions in One-shot Federated Learning. Numbers refer to the subsection in Section \ref{sec:Open} where each challenge is elaborated. }
    \label{fig:future}
\end{figure}
Despite notable advancements, One-Shot Federated Learning (One-Shot FL) faces unresolved challenges and research gaps that impede its practical deployment. Current efforts focus on enhancing accuracy through techniques such as model ensembles, model distillation, and data distillation. However, these approaches often risk compromising the privacy of client models and local data, as depicted in Figure \ref{fig:future}. In this section, this survey outlines critical challenges, research gaps, and future directions to drive progress in One-Shot FL.
\subsection{6.1 Challenges}
Current research focuses on studying model ensembles or model distillation and data distillation methods to improve the accuracy of One-shot Federated Learning. However, from this perspective, these methods also pose challenges to the privacy of client models and their private data. This survey profoundly summarizes the current challenges of One-shot Federated Learning.
Therefore, looking forward, this survey provides some perspectives for future work, such as studying aggregator models that do not require access to all client models in the ensemble.

\subsubsection{6.1.1 Data Access Minimization}
One of the most pressing challenges facing One-shot Federated Learning (One-shot FL) is the extremely limited access to data, a limitation further exacerbated by its reliance on a single round of communication. Unlike traditional federated learning, which achieves iterative model optimization and verification through multiple rounds of communication, One-shot FL must complete model convergence and optimization in a single interaction. This makes it critical to efficiently utilize on-device data, which is often sparse, heterogeneous, and non-independent and identically distributed (Non-IID). In addition, the lack of centralized or proxy datasets severely limits the ability to verify, audit, or fine-tune models, raising widespread concerns about their reliability and generalization in real-world applications.

This limitation has given rise to a series of key research questions: How to design auditing or verification techniques that can operate effectively with only limited device data in a single round of training? Traditional verification methods often rely on centralized datasets or multiple rounds of testing, which are incompatible with the single-round communication constraint of One-shot FL. Therefore, it is urgent to develop new methods to ensure rigorous evaluation of models without compromising system privacy or efficiency. In addition, can synthetic data generation or data augmentation techniques be adapted to One-shot FL to improve the robustness of the model without violating privacy constraints? For example, techniques such as data synthesis based on differential privacy or federated generative adversarial networks (GANs) can improve model performance by generating representative data samples while protecting the privacy of local datasets.

Addressing these challenges is critical to ensuring the reliability and practicality of One-shot FL models in real-world applications. For example, in the medical field, patient data are highly sensitive and decentralized; in IoT scenarios, device resources are limited and data are unevenly distributed. In these scenarios, how to achieve model verification and optimization with limited data access is an urgent problem to be solved. Current research has begun to explore directions such as federated learning benchmarks (such as NoIID-Bench, FLamby) and privacy-preserving data generation, but there are still significant research gaps. Future research should focus on developing lightweight verification frameworks, privacy-aware data augmentation techniques, and robust aggregation methods to achieve efficient and privacy-preserving model training under the strict constraints of One-shot FL. By filling these gaps, One-shot FL is expected to realize its great potential in resource-constrained and privacy-sensitive scenarios.
\begin{figure}[htbp]
    \centering
    \includegraphics[width=\linewidth]{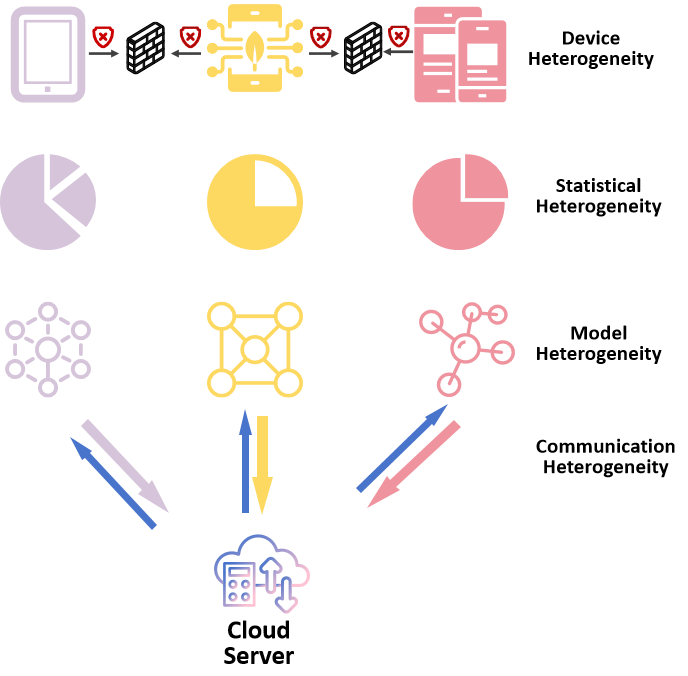}
    \caption{Schematic diagram of heterogeneous One-shot Federated Learning, including data heterogeneity, device heterogeneity, model heterogeneity, and communication heterogeneity}
    \label{fig:enter-label}
\end{figure}
\subsubsection{6.1.2 Trustworthies}
In practical applications, a core challenge faced by One-shot Federated Learning (One-shot FL) is data privacy and security. Since One-shot FL completes model training only in a single round of communication, participants share model parameters or gradient information in one interaction. Although this information does not directly expose the original data, it may still be used by attackers to infer sensitive information. For example, attackers can reconstruct the participants' local training data from the shared model parameters through model inversion attack or gradient inversion attack, thereby threatening data privacy. This attack is particularly dangerous in highly sensitive application scenarios such as medical and financial, because the data in these fields often contain personal privacy or commercial secrets.

Although One-shot FL reduces the privacy risks brought by multiple communications in traditional federated learning by reducing the number of communication rounds, its single-round communication characteristics also introduce new challenges. First, the limitation of single-round communication makes the design of privacy protection technology more complicated. Traditional differential privacy or secure aggregation protocols usually rely on multiple rounds of interactions to balance privacy and model performance, while in One-shot FL, these techniques must complete privacy protection within a single round while ensuring the practicality of the model. Second, the heterogeneity of data distribution further exacerbates the risk of privacy leakage. Since the data of participants are usually non-independent and identically distributed (Non-IID), it may be easier for attackers to infer the data characteristics of specific participants from the model parameters.

To address these challenges, an important research direction is to develop lightweight privacy protection mechanisms that can operate effectively in a single round of communication. For example, noise addition techniques based on differential privacy can perturb local model parameters before model aggregation, thereby preventing data reconstruction attacks. In addition, the optimization of secure aggregation protocols is also a key direction, especially when dealing with large-scale participants, how to achieve efficient privacy protection without significantly increasing computational overhead. However, existing secure aggregation methods usually rely on complex mask generation and reconstruction mechanisms, which may be difficult to implement on resource-constrained devices (such as IoT devices). Therefore, future research needs to explore more efficient and lightweight privacy protection schemes to achieve a balance between privacy and performance in One-shot FL.

In short, One-shot FL faces severe challenges in data privacy and security. Especially under the limitation of single-round communication, how to protect the privacy of participants' data without sacrificing model performance is still an urgent problem to be solved. By combining differential privacy, secure aggregation and lightweight computing technology, future research is expected to provide One-shot FL with more reliable security guarantees, thereby promoting its widespread application in privacy-sensitive fields.
\subsubsection{6.1.3 Scalability and Resource Constraints}
The rapid growth of Internet of Things (IoT) devices and edge networks introduces challenges in Federated Learning (FL) due to the resource limitations of edge devices, such as limited computational power, memory, and bandwidth~\cite{imteaj2021survey}. These constraints necessitate the design of efficient algorithms capable of operating under stringent conditions.

Two key research questions arise in this survey. First, how can lightweight algorithms for model aggregation, privacy preservation, and performance evaluation be developed to suit the limited capabilities of resource-constrained devices? Traditional model aggregation in FL is computationally intensive, and incorporating privacy-preserving techniques further increases both computational and communication overheads. Developing methods that minimize these costs while maintaining model accuracy and privacy is essential.

Second, can communication-efficient techniques be devised to operate within the constraints of a single communication round while preserving model accuracy and ensuring privacy? One-shot Federated Learning (One-shot FL) addresses communication overhead by limiting interactions to a single round, but this presents challenges in preserving model performance and privacy in such a constrained setting. Efficient aggregation and privacy-preserving mechanisms that do not compromise accuracy are crucial for its practical implementation.

Addressing these challenges is vital for enabling the scalable and practical deployment of FL in resource-constrained environments, where both efficiency and performance are critical. Future work should explore hybrid approaches combining One-shot FL with multi-round techniques, leveraging model compression, quantization, and communication-efficient algorithms. Additionally, novel privacy-preserving methods tailored for edge devices need to be developed to ensure both security and scalability.
\subsubsection{6.1.4 Privacy-Accuracy Trade-offs}

One-shot Federated Learning (One-shot FL) poses a more difficult challenge in balancing the trade-off between model accuracy and privacy protection in a single communication. Traditional federated learning methods usually rely on multiple rounds of communication to iteratively optimize model parameters and achieve the best balance between these two objectives. However, the fundamental limitation of One-shot Federated Learning is confined to a single round of communication, which complicates the task of achieving both high accuracy and strong privacy guarantees simultaneously.

The privacy-accuracy trade-off is a well-known challenge in machine learning and is particularly important in decentralized settings such as FL, where data privacy is critical. To address this issue in the context of One-shot Federated Learning, it is imperative to explore new aggregation and regularization techniques that can effectively preserve privacy while maintaining competitive model accuracy. Traditional techniques such as differential privacy (DP) and secure aggregation can enhance privacy but often introduce noise or computational overhead that degrades model performance. In the context of One-shot Federated Learning, the introduction of such methods in a single round must be carefully balanced to prevent significant accuracy loss. Communication efficiency is another critical concern, particularly for resource-constrained edge devices. Techniques like model pruning and quantization reduce communication overhead by transmitting only critical parameters or low-bit representations. However, aggressive compression can irreversibly discard information vital for model convergence. For instance, in a federated language modeling task,~\cite{sattler2021fedaux} found that pruning 80\% of model weights in FL led to a 22\% drop in perplexity scores compared to uncompressed baselines. Similarly, knowledge distillation—where clients train lightweight "student" models—can reduce communication costs but risks losing nuanced patterns captured by larger models.

A key research question is: How to optimize the trade-off between privacy and accuracy within the limitation of a single round of communication? Achieving this goal will require innovations in algorithmic design and theoretical foundations. For example, can we leverage privacy-preserving aggregation methods (such as homomorphic encryption or secure multi-party computation) without sacrificing model utility? Furthermore, can regularization techniques (such as model pruning or noise injection) be selectively applied to ensure that privacy is preserved while still allowing efficient learning in a single round of communication?  For instance, adding calibrated noise to local model updates (e.g., Gaussian or Laplacian noise) can protect against membership inference attacks by obscuring sensitive data patterns. However, excessive noise injection in a single round risks severely degrading model utility.~\cite{mendietaexploring} demonstrated that in image classification tasks on CIFAR-10, applying DP to One-shot FL reduced accuracy by 12-18\% compared to non-private baselines, highlighting the steep cost of privacy guarantees in this setting. 
\subsection{6.2 Research Gaps}
The unique constraints of One-shot Federated Learning (One-shot FL) exacerbate several challenges that are already present in traditional federated learning (FL). 
These challenges create significant research gaps that need to be addressed to enable the practical deployment of One-shot FL in real-world applications. Below, this survey outlines these gaps in the context of One-shot FL:
\begin{itemize}
    \item One-shot FL operates under the constraint of a single communication round, making it even more critical to leverage on-device data effectively. However, the lack of access to centralized or proxy data limits the ability to validate or audit models.
    \item Privacy can be further enhanced in various One-shot FL algorithms such as FENS~\cite{allouah2024revisiting} through techniques such as differential privacy~\cite{geyer2017differentially} or trusted execution environments~\cite{messaoud2022shielding}. 
\end{itemize}

\subsection{6.3 Future Directions}
In this section, this survey discuss the emerging trends and potential future research directions in One-shot Federated Learning.

\subsubsection{6.3.1 Integrate IoT Device with One-shot FL}
Existing object recognition technologies for IoT and self-driving cars are far from perfect. The main drawback of these algorithms is that they cannot handle the large amount of unlabeled data faced in the real world and can only operate with limited data~\cite{arisdakessian2022survey}. In future research, consider integrating state-of-the-art technologies such as One-shot Federated Learning and reinforcement learning to study strategies that can enhance object detection in IoT, such as studying the connectivity and integration schemes of these technologies for object detection in IoT environments.
\subsubsection{6.3.2 Integrate Satellite Communication with One-shot FL} 
In terms of applications, the application of One-shot Federated Learning will include exploring various LEO constellations, ranging from sparse to dense constellations, GS located in different geographical locations, and training heterogeneous ML models across satellites and constellations.
\subsubsection{6.3.3 Integrate Scientific Machine Learning with One-shot FL}
Scientific Machine Learning (SciML) designs machine learning methods that predict physical systems governed by partial differential equations (PDE). 
SciML integrates physics-based models with data-driven machine learning techniques to improve model interpretability and accuracy in domains such as climate modeling and material science.
Physically-informed neural networks (PINNs) can efficiently solve ODE/PDE by embedding physical laws into neural networks. The core goal of SciML is to embed physical laws (such as ODE/PDE) into machine learning models (such as PINNs) to ensure that the predictions of the models conform to physical laws~\cite{zhang2024federatedscientific}. One-shot Federated Learning (One-shot FL) is an efficient Federated Learning variant in which the participants only need one communication to complete model aggregation, which is suitable for scenarios with high communication costs or strict data privacy requirements.

Therefore, in One-shot Federated Learning, the participants can make up for data shortage or heterogeneity problems by generating synthetic data (such as simulation data based on physical equations). The characteristic of SciML is that SciML usually involves high-dimensional data (such as PDE solving in three-dimensional space) and complex models (such as deep neural networks), which have high computational costs. Therefore, One-shot Federated Learning can provide SciML with an efficient and privacy-preserving distributed computing framework, which is particularly suitable for large-scale scientific computing tasks.

The potential value of fusion is that it can design federated optimization algorithms that combine physical constraints to ensure that the model effectively integrates physical information and scientific data in One-shot Federated Learning to meet the laws of physics.

For specific applications, multiple participants have physical simulation data in different regions and use One-shot Federated Learning to collaboratively train a global PINNs model to solve PDEs (such as fluid dynamics equations). In the medical or defense fields, One-shot Federated Learning is used to train PINNs models to ensure that sensitive data are not leaked. The fusion of Scientific Machine Learning (SciML) and One-shot Federated Learning (One-shot FL) is a very promising research direction. The distributed data requirements of SciML and the distributed learning capabilities of One-shot Federated Learning are naturally compatible, making their respective characteristics and advantages complementary, and can efficiently utilize decentralized scientific data to solve many problems that are difficult to deal with traditional methods.
\subsubsection{6.3.4 Recommend Benchmark}
One-shot Federated Learning (One-shot FL) has emerged as a promising paradigm for efficient and privacy-preserving distributed machine learning, enabling collaborative model training with minimal communication overhead. However, the lack of standardized benchmarks and evaluation protocols has hindered the fair comparison of methods and the identification of best practices in this field. 
This paper presents a feasible and comprehensive benchmarking framework for One-shot FL that covers datasets, evaluation metrics, and challenges specific to this learning paradigm to address this gap.

Due to the communication efficiency, data heterogeneity, and privacy constraints of One-shot FL, benchmarking One-shot FL faces three major challenges: (1) communication constraints, (2)data heterogeneity, (3) privacy and security.

From the perspective of ethical consideration, future works can address biases and fairness in federated learning systems.
\paragraph{Datasets}
To facilitate fair and comprehensive evaluation, we propose a diverse set of benchmark datasets tailored to One-shot FL, including Image Classification, Healthcare, Scientific Computing, IoT, and Edge Computing. Through diverse datasets, the practicality and effectiveness of One-shot Federated Learning methods can be more comprehensively evaluated.

\paragraph{Evaluation Metrics}
We define a set of evaluation metrics to assess the performance, efficiency, and privacy of One-shot FL methods:
\begin{itemize}
    \item Model Accuracy: Standard metrics like accuracy, F1-score, and mean squared error (MSE) to evaluate model performance.

    \item Communication Efficiency: Metrics such as communication cost (e.g., bytes transmitted) and latency to measure the efficiency of the single communication round.

    \item Privacy and Security: Metrics like privacy loss and robustness to adversarial attacks(ASR).

    \item Scalability: Metrics to evaluate the performance of One-shot FL as the number of clients increases (e.g., convergence speed, resource usage).
\end{itemize}
\paragraph{Modular Framework}
The modular benchmarking framework provides the following features for research and application:
\begin{itemize}
    \item (1) Simulate various scenarios: configure datasets, communication constraints, and privacy settings to simulate real-world conditions.
    
\item (2) Compare methods: evaluate state-of-the-art One-shot FL methods (e.g., model averaging, knowledge distillation) against baseline methods.

\item (3) Analyze trade-offs: study the trade-offs between model performance, communication efficiency, and privacy.
\end{itemize}
\section{7 Conclusion}
In this paper, we examine the approaches, inherent challenges, and future directions of One-shot Federated Learning (One-shot FL), a transformative paradigm in distributed machine learning. By enabling collaborative model training with minimal communication overhead and robust privacy guarantees, One-shot FL has emerged as a promising solution for applications spanning healthcare, scientific computing, and beyond. Yet, its practical implementation and scalability are fraught with complexities that demand innovative solutions and interdisciplinary collaboration.

A central challenge in One-shot Federated Learning (One-shot FL) is data heterogeneity, as non-IID datasets hinder model performance. Advanced aggregation techniques and personalized learning strategies are needed to balance global accuracy with local adaptability. Communication efficiency is another critical issue, which is addressed through methods like model distillation and gradient sparsification. Privacy preservation relies on cryptographic methods such as secure multi-party computation (MPC) and differential privacy(DP), though these often introduce trade-offs in computational overhead and model utility.

The future of One-shot FL lies in cross-disciplinary innovation, integrating cryptography, machine learning, and distributed systems. Combining physics-informed neural networks (PINNs) with One-shot FL could enable high-fidelity modeling, while energy-efficient strategies reduce environmental costs. Standardized benchmarks and evaluation frameworks are essential for reproducibility and progress.

In summary, One-shot FL offers a transformative balance of efficiency, privacy, and scalability. There are still many challenges and future directions to be explored in this field.
\section{CRediT authorship contribution statement}
\textbf{Flora Amato}:
Validation, Formal analysis, Conceptualization

\textbf{Lingyu Qiu}:
Investigation, Resources, Methodology, Formal analysis,
Visualization, Writing – original draft.

\textbf{Mohameed Tanveer}:
Supervision, Validation, Conceptualization

\textbf{Salvatore Cuomo}:
Validation, Conceptualization

\textbf{Fabio Giampaolo}:
Validation, Conceptualization, Methodology

\textbf{Francesco Piccialli}:
Supervision, Methodology, Validation, Conceptualization

\subsection{Declaration of Competing Interest}
The authors declare that they have no known competing financial interests or personal relationships that could have appeared to
influence the work reported in this paper.

\subsection{Data Availability}

No data was used for the research described in the article.

\newpage
\bibliographystyle{cas-model2-names}
\bibliography{aaai25}

\end{document}